\documentclass[pdflatex,sn-mathphys-num]{sn-jnl}% Math and Physical Sciences Numbered Reference Style
\usepackage{graphicx}%
\usepackage{multirow}%
\usepackage{amsmath,amssymb,amsfonts}%
\usepackage{amsthm}%
\usepackage{mathrsfs}%
\usepackage[title]{appendix}%
\usepackage{xcolor}%
\usepackage{textcomp}%
\usepackage{manyfoot}%
\usepackage{booktabs}%
\usepackage{algorithm}%
\usepackage{algorithmicx}%
\usepackage{algpseudocode}%
\usepackage{listings}%
\theoremstyle{thmstyleone}%
\theoremstyle{thmstyletwo}%

\theoremstyle{thmstylethree}%

\usepackage{pdflscape}
\usepackage{tabularx}
\usepackage{booktabs}
\usepackage{multirow}

\begin{document}

% \title[Article Title]{Article Title}
% \title[Article Title]{Advances in AI-Powered Quantitative Remote Sensing of the Terrestrial Surface: Methods, Models, and Applications}
\title[Article Title]{From Physics to Foundation Models: A Review of AI-Driven Quantitative Remote Sensing Inversion}

%%=============================================================%%
%% GivenName	-> \fnm{Joergen W.}
%% Particle	-> \spfx{van der} -> surname prefix
%% FamilyName	-> \sur{Ploeg}
%% Suffix	-> \sfx{IV}
%% \author*[1,2]{\fnm{Joergen W.} \spfx{van der} \sur{Ploeg} 
%%  \sfx{IV}}\email{iauthor@gmail.com}
%%=============================================================%%

% \author*[1,2]{\fnm{First} \sur{Author}}\email{iauthor@gmail.com}

% \author[2,3]{\fnm{Second} \sur{Author}}\email{iiauthor@gmail.com}
% \equalcont{These authors contributed equally to this work.}

% \author[1,2]{\fnm{Third} \sur{Author}}\email{iiiauthor@gmail.com}
% \equalcont{These authors contributed equally to this work.}

% \affil*[1]{\orgdiv{Department}, \orgname{Organization}, \orgaddress{\street{Street}, \city{City}, \postcode{100190}, \state{State}, \country{Country}}}

% \affil[2]{\orgdiv{Department}, \orgname{Organization}, \orgaddress{\street{Street}, \city{City}, \postcode{10587}, \state{State}, \country{Country}}}

% \affil[3]{\orgdiv{Department}, \orgname{Organization}, \orgaddress{\street{Street}, \city{City}, \postcode{610101}, \state{State}, \country{Country}}}

\author[1,2,3]{\fnm{Zhenyu} \sur{Yu}}\email{yuzhenyuyxl@foxmail.com}

\author[3]{\fnm{Mohd Yamani Idna} \sur{Idris}}\email{yamani@um.edu.my}

\author*[2]{\fnm{Hua} \sur{Wang}}\email{wangh\_df@163.com}

\author*[1]{\fnm{Pei} \sur{Wang}}\email{peiwang@kust.edu.cn}

\author[4]{\fnm{Junyi} \sur{Chen}}\email{joel59cjy@163.com}

\author[5]{\fnm{Kun} \sur{Wang}}\email{wk@um.edu.my}

% \author[6]{\fnm{Qi} \sur{Wang}}\email{qiwang@sicau.edu.cn}

% \author[7]{\fnm{Fei} \sur{Ma}}\email{mafei@gml.edu.cn}

% \author[8]{\fnm{Wenbin} \sur{Zhang}}\email{wenbinzhang2008@gmail.com}

% \author[9]{\fnm{Rizwan} \sur{Qureshi}}\email{engr.rizwanqureshi786@gmail.com}

% \author[6]{\fnm{Xiang} \sur{Yong}}\email{yong.xiang@deakin.edu.au}

% \affil[1]{\orgdiv{School of Information Science and Technology}, \orgname{Yunnan Normal University}, \orgaddress{\city{Kunming}, \postcode{650000}, \country{China}}}

\affil[1]{\orgdiv{Faculty of Information Engineering and Automation}, \orgname{Kunming University of Science and Technology}, \orgaddress{\city{Kunming}, \postcode{650093}, \country{China}}}

\affil[2]{\orgdiv{School of Business Administration}, \orgname{Zhejiang University of Finance and Economics Dongfang College}, \orgaddress{\city{Jiaxing}, \postcode{310018}, \country{China}}}
% \affil[3]{\orgdiv{School of Earth Sciences and Engineering}, \orgname{Hohai University}, \orgaddress{\city{Nanjing}, \postcode{210098}, \country{China}}}

\affil[3]{\orgdiv{Faculty of Computer Science and Information Technology}, \orgname{Universiti Malaya}, \orgaddress{\city{Kuala Lumpur}, \postcode{50603}, \country{Malaysia}}}

\affil[4]{\orgdiv{Faculty of Land Resources Engineering}, \orgname{Kunming University of Science and Technology}, \orgaddress{\city{Kunming}, \postcode{650093}, \country{China}}}

\affil[5]{\orgdiv{School of Information}, \orgname{Yunnan University of Finance and Economic}, \orgaddress{\city{Kunming}, \postcode{650221}, \country{China}}}

% \affil[6]{\orgdiv{College of Resources}, \orgname{Sichuan Agricultural University}, \orgaddress{\city{Chengdu}, \postcode{611130}, \country{China}}}

% \affil[7]{\orgname{Guangming Laboratory}, \orgaddress{\city{Shenzhen}, \postcode{518060}, \country{China}}}

% \affil[8]{\orgdiv{Knight Foundation School of Computing and Information Sciences}, \orgname{Florida International University}, \orgaddress{\city{Miami}, \postcode{33199}, \country{USA}}}

% \affil[9]{\orgdiv{Center for research in Computer Vision}, \orgname{University of Central Florida}, \orgaddress{\city{Orlando}, \postcode{32816}, \country{USA}}}

% \affil[6]{\orgdiv{Faculty of Science Engineering and Built Environment}, \orgname{Deakin University}, \orgaddress{\city{Burwood}, \postcode{3125}, \country{Australia}}}

% \affil[*]{\orgdiv{Hua Wang (wangh\_df@163.com) and Pei Wang (peiwang@kust.edu.cn)}}

% \affil[8]{\orgdiv{xxxxx}, \orgname{xxxxxx}, \orgaddress{\city{xxxxx}, \postcode{xxxxx}, \country{xxx}}}

%%==================================%%
%% Sample for unstructured abstract %%
%%==================================%%

\abstract{
Quantitative remote sensing inversion aims to estimate continuous surface variables—such as biomass, vegetation indices, and evapotranspiration—from satellite observations, supporting applications in ecosystem monitoring, carbon accounting, and land management. With the evolution of remote sensing systems and artificial intelligence, traditional physics-based paradigms are giving way to data-driven and foundation model (FM)–based approaches. This paper systematically reviews the methodological evolution of inversion techniques, from physical models (e.g., PROSPECT, SCOPE, DART) to machine learning methods (e.g., deep learning, multimodal fusion), and further to foundation models (e.g., SatMAE, GFM, mmEarth). We compare the modeling assumptions, application scenarios, and limitations of each paradigm, with emphasis on recent FM advances in self-supervised pretraining, multi-modal integration, and cross-task adaptation. We also highlight persistent challenges in physical interpretability, domain generalization, limited supervision, and uncertainty quantification. Finally, we envision the development of next-generation foundation models for remote sensing inversion, emphasizing unified modeling capacity, cross-domain generalization, and physical interpretability.
}


%%\pacs[JEL Classification]{D8, H51}

%%\pacs[MSC Classification]{35A01, 65L10, 65L12, 65L20, 65L70}

\maketitle

\section{Introduction}
\textbf{Quantitative remote sensing inversion} serves as a critical link between remote observations and geophysical parameter estimation, underpinning key applications such as land use change monitoring, ecosystem evaluation, water resource management, and carbon cycle modeling. With the advancement of remote sensing systems from single optical sensors to multi-source and heterogeneous platforms—including multispectral (MSI) \cite{shu2023heterodimensional}, hyperspectral (HSI) \cite{sun2024applications}, synthetic aperture radar (SAR) \cite{hashemi2024review}, and LiDAR—remote sensing data \cite{min2024incorporating} now exhibit high spatiotemporal resolution and multimodal characteristics, offering a solid foundation for the accurate inversion of surface state variables \cite{wulder2022fifty, karim2023current, li2023big, han2023survey, qianqian2022research}. Traditional inversion approaches based on physical models \cite{abbes2024advances} and look-up tables \cite{wang2024exploring} ensure strong physical consistency \cite{chen2024digital} and interpretability \cite{guo2024novel}, yet they are often limited by modeling complexity, parameter accessibility, and poor generalization. Data-driven methods, such as random forests (RF) \cite{tan2024estimating}, support vector regression (SVR) \cite{chen2024ensemble}, and deep neural networks \cite{yasir2024shipgeonet}, offer improved flexibility and predictive accuracy but suffer from dependence on high-quality labeled data and lack inherent physical constraints \cite{bai2024integrating}. Recently, foundation models (FMs) \cite{hong2024spectralgpt}, characterized by large-scale self-supervised pretraining, multi-task learning, and cross-modal fusion, have gained prominence. Representative models such as SatMAE \cite{cong2022satmae}, GFM \cite{mendieta2023towards}, and MMEarth \cite{nedungadi2024mmearth} demonstrate notable advantages in representation learning, transferability, and scientific reasoning \cite{bommasani2021opportunities, nedungadi2024mmearth}, paving the way toward a paradigm shift from task-specific customization to unified, general-purpose architectures. Thus, a systematic review of the current progress, modeling principles, and future trends of FMs in quantitative remote sensing inversion is essential for advancing theoretical understanding and guiding the development of robust and scalable inversion frameworks.

\begin{figure}
    \centering
    \includegraphics[width=1\linewidth]{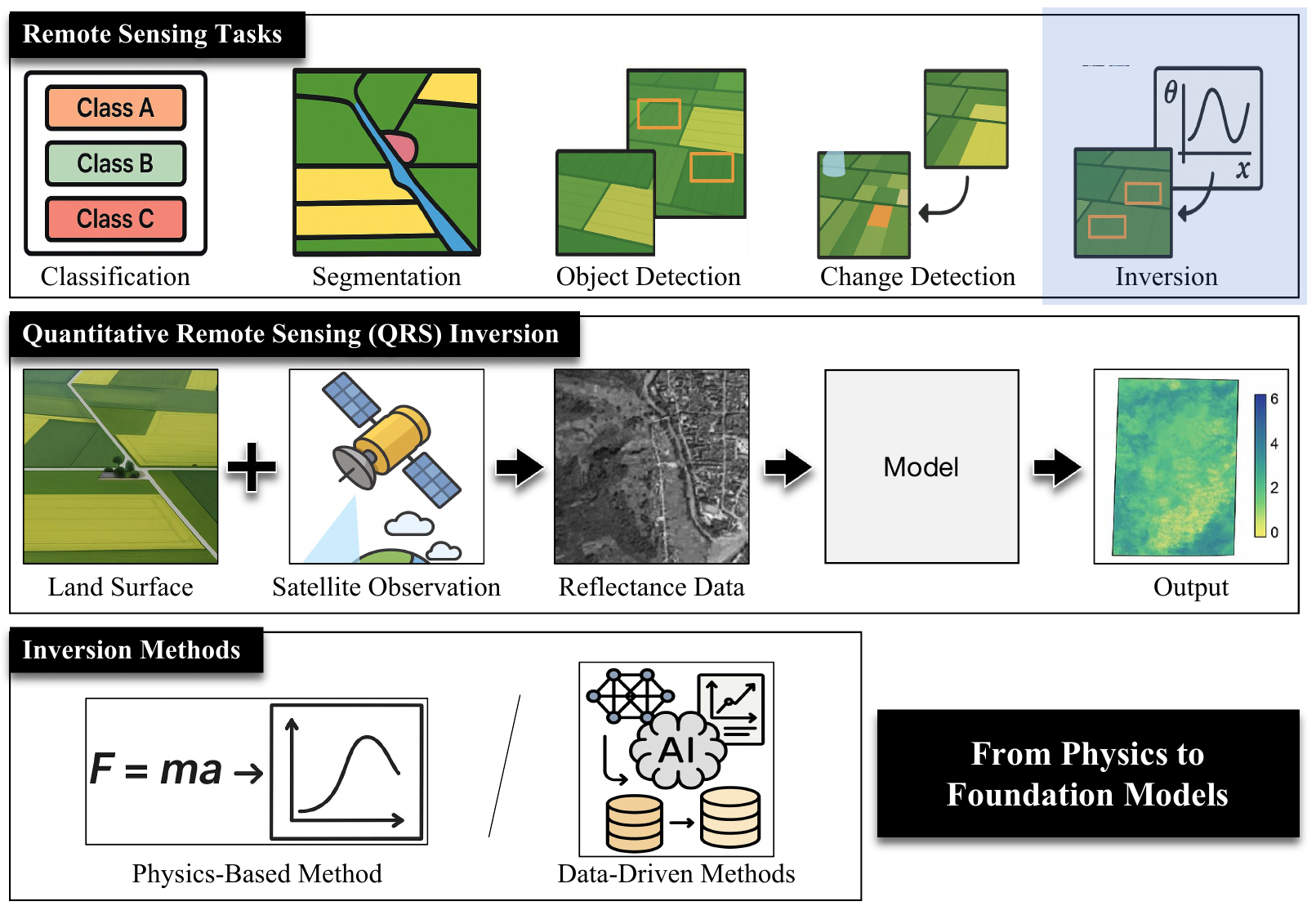}
    \caption{Overview of representative remote sensing inversion tasks, including typical input modalities, output variables, and application scenarios.}
    \label{fig_overview}
\end{figure}

\textbf{Common Inversion Tasks and Output Characteristics.} Quantitative remote sensing inversion focuses on the high-precision estimation of continuous geophysical parameters, which serve as critical inputs for environmental modeling and geoscientific analysis \cite{yu2025qrs,yu2025ai}. These variables hold broad scientific and practical value. As illustrated in Figure~\ref{fig_overview}, quantitative inversion aims to derive a range of physically meaningful surface parameters from multi-source observational data, including leaf area index (LAI) \cite{li2024chlorophyll}, normalized difference vegetation index (NDVI) \cite{almouctar2024drought}, fraction of absorbed photosynthetically active radiation (FAPAR) \cite{mallick2024net}, aboveground biomass (AGB) \cite{ma2024development}, and canopy height \cite{malambo2024mapping}—all of which provide essential foundations for ecological monitoring and carbon cycle modeling \cite{guo2021mapping}. Additional parameters such as evapotranspiration (ET) \cite{tang2024spatial}, soil moisture (SM) \cite{mane2024advancements}, surface albedo \cite{chen2024surface}, surface roughness \cite{djurovic2024modeling}, total suspended solids (TSS) \cite{harringmeyer2024hyperspectral}, and turbidity \cite{grandjean2024critical} are widely used in hydrological process simulation and climate response analysis \cite{bai2024integrating,yu2025rainy}. These variables collectively reflect the structural features of terrestrial ecosystems, land-atmosphere energy and water exchange processes, and carbon–nitrogen cycling characteristics, serving as indispensable inputs to Earth system science \cite{han2023survey,yu2024iidm}. In recent years, improvements in the spatial, spectral, and temporal capabilities of remote sensing platforms have led to high-resolution, multimodal, and high-frequency observation systems, enabling inversion tasks to become increasingly high-dimensional, multi-scale, and heterogeneous \cite{yue2025diffusion,wang2025lmfnet}. This evolution imposes growing demands on modeling techniques in terms of feature representation, cross-regional generalization, and uncertainty quantification \cite{lu2025uncertainty}. Consequently, the accurate and robust inversion of such geophysical variables has emerged as a core challenge in advancing remote sensing from perception-driven observations toward cognitively informed understanding.

\textbf{Evolution of Methodological Paradigms.} The development of quantitative remote sensing inversion methods has progressed from physically based modeling approaches to data-driven machine learning frameworks, and more recently, toward the emerging paradigm of foundation models (FMs) that aim to provide general-purpose learning capabilities \cite{karim2023current, bai2024integrating,yu2025satellitemaker}. Early physical models, such as PROSAIL, SCOPE, and DART, are grounded in radiative transfer theory and process-based simulations, establishing forward models that link remote sensing observations to surface parameters. These methods emphasize physical interpretability and consistency \cite{verrelst2015optical,yu2025satellitecalculator}. However, when faced with multi-source heterogeneous data, high-dimensional nonlinear relationships, and large-scale automation demands, these approaches suffer from limitations in parameter accessibility and modeling flexibility. In the data-driven era, machine learning and deep learning methods—such as RF, SVR, CNNs, GNNs, and Transformers—have demonstrated strong modeling capabilities. By enabling end-to-end learning of nonlinear mappings between remote sensing observations and geophysical variables, these methods significantly improve predictive accuracy and generalization \cite{han2023survey, lei2021transformer}. Nonetheless, they often require large volumes of labeled data and lack intrinsic physical constraints, which limits their transferability, scientific interpretability, and controllability across diverse regions \cite{luo2025steam}. In recent years, the paradigm of foundation models has gained traction, characterized by self-supervised pretraining, multi-task adaptation, and multimodal fusion. These models aim to construct universal representation frameworks for remote sensing, exhibiting transformative potential in weakly supervised learning, cross-modal perception, and geoscientific reasoning \cite{bommasani2021opportunities, nedungadi2024mmearth}. This trend marks a significant shift from task-specific modeling approaches toward structurally unified paradigms in remote sensing inversion.

To systematically review the methodological evolution of quantitative remote sensing inversion, this study categorizes existing technical approaches into \textbf{three representative paradigms}: physics-based models, data-driven methods, and foundation model frameworks. The first category, physics-based models, relies on radiative transfer mechanisms and surface process simulations. Representative models such as PROSAIL \cite{zhang2024monitoring} and SCOPE \cite{ntakos2024coupled} establish physical mappings between observed radiance and geophysical variables, emphasizing theoretical interpretability and scientific consistency. These methods are well-suited for inversion scenarios with clearly defined physical mechanisms and strong prior constraints \cite{verrelst2015optical, mutanga2023spectral,yu2025dc4cr}. The second category, data-driven methods, builds upon statistical learning and deep neural networks to directly model nonlinear relationships from remote sensing observations. These approaches exhibit flexible model expression capabilities and high predictive accuracy, showing significant advances in multi-source data fusion and automated modeling \cite{han2023survey, bai2024integrating,yu2025satelliteformula}. However, their limitations in regional generalization, scientific consistency, and reliance on labeled samples remain unresolved. The third paradigm, foundation models, features large-scale self-supervised pretraining, unified representation learning, and cross-modal task adaptation. Representative models such as SatMAE, GFM, and MMEarth aim to enhance adaptability and transferability in multi-task, multi-modal, and cross-regional remote sensing applications by adopting unified architectures and general-purpose task interfaces \cite{bommasani2021opportunities, nedungadi2024mmearth}. This review focuses on quantitative inversion tasks for surface parameters (e.g., NDVI, LAI, AGB, SIF), systematically analyzing the modeling mechanisms, adaptation characteristics, and methodological advantages of the three paradigms. By clarifying their technical boundaries and evolutionary trends, this work seeks to provide a theoretical foundation and methodological guidance for constructing the next generation of remote sensing inversion frameworks with high adaptability and generalization capacity.

Although artificial intelligence technologies have been widely applied to remote sensing tasks such as image classification, change detection, and scene understanding in recent years—with a growing number of relevant review articles \cite{bai2024integrating, han2023survey}—a systematic review specifically focused on quantitative remote sensing inversion, particularly under the emerging foundation model paradigm, remains limited. At present, a comprehensive understanding of how foundation models are constructed, adapted, and evolved for inversion tasks is still lacking. This has led to gaps in guiding researchers on model selection, paradigm integration, and multimodal modeling strategies. This review attempts to address this gap by offering a preliminary systematic synthesis, with the following \textbf{contributions}:

\begin{itemize}
    \item \textbf{Task-centered perspective:} This paper focuses on a core challenge in remote sensing—quantitative inversion of surface parameters—and analyzes the modeling characteristics and inversion difficulties of typical geophysical variables such as LAI, NDVI, AGB, and ET. It clarifies both the unique and generalizable aspects of applying AI models to these tasks.

    \item \textbf{Paradigm evolution framework:} We propose a three-stage evolutionary framework, transitioning from physics-based modeling to data-driven methods, and finally to foundation models. This structure summarizes the theoretical underpinnings, modeling strategies, and application scopes of each paradigm, highlighting key turning points.

    % \item \textbf{Challenges and future directions:} We discuss major open challenges in current models, including scientific consistency, small-sample generalization, uncertainty quantification, and regional adaptability. We further envision the potential of “Remote Sensing GPT” in enabling multi-task fusion, cross-modal perception, and knowledge-guided modeling.
    \item \textbf{Toward unified inversion frameworks}: We outline future directions for foundation model–driven inversion systems with multi-task, multimodal, and cross-domain adaptability, paving the way toward scalable, interpretable, and general-purpose modeling in quantitative remote sensing.
\end{itemize}

% 遥感反演任务示例图
% fig_overview

% 分类体系图
% fig_catogaries
\begin{figure}
    \centering
    \includegraphics[width=0.7\linewidth]{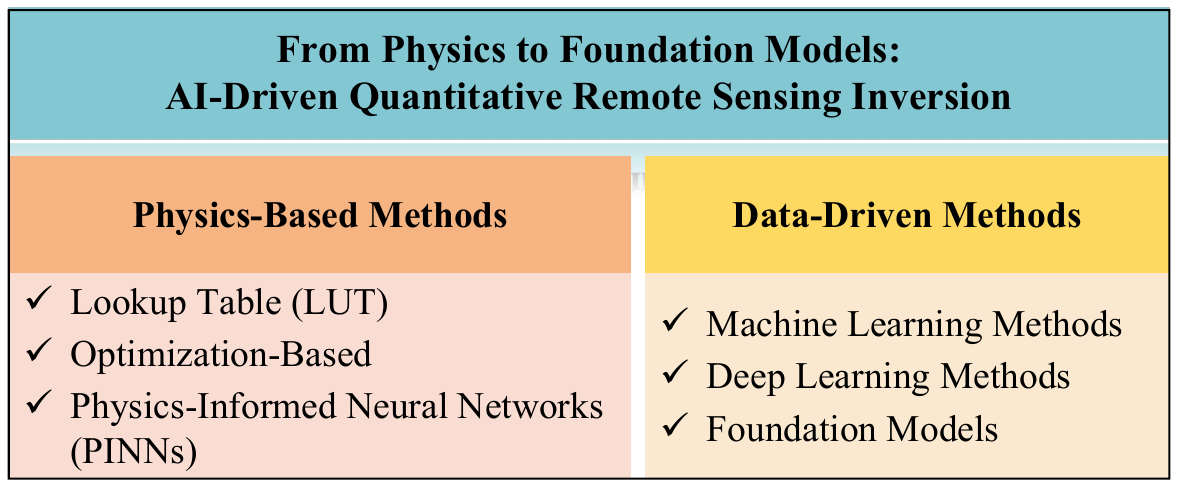}
    \caption{Taxonomy of remote sensing inversion methods, categorized into physics-based methods, data-driven methods, and hybrid/physics-informed methods.}
    \label{fig_catogaries}
\end{figure}

\section{Task Taxonomy and Attributes}
\subsection{Task Definition}
% \section{Task Definition: Quantitative Remote Sensing Inversion}

\textbf{Quantitative Remote Sensing Inversion} refers to the process of deriving a set of geophysical parameters with explicit physical meanings from multi-source observational data acquired by remote sensing platforms. This task plays a central role in a wide range of critical applications, including ecosystem structure assessment, carbon stock estimation, hydrological modeling, and surface energy exchange studies. Unlike discriminative tasks such as classification or detection, quantitative inversion focuses on the numerical prediction of continuous variables that describe the structure, function, or state of the Earth’s surface. These results are expected not only to achieve pixel-level spatial accuracy but also to ensure regional consistency and process-level interpretability \cite{hu2023physical, li2023big}.

A key challenge in this task lies in the fact that remote sensing observations typically capture radiative or reflective signals (e.g., spectral reflectance, radar backscatter, or LiDAR return intensity), while the target variables are often physical quantities of Earth surface processes (e.g., LAI, NDVI, AGB, ET, SM). There is no simple linear or explicit functional mapping between them. The inversion process can typically be formulated as a functional learning problem:

\begin{equation}
\hat{y} = f(\mathbf{x}) \quad \text{or} \quad \hat{y} = \arg\min_{\theta} \mathcal{L}(f_{\theta}(\mathbf{x}), y)
\end{equation}

where $\mathbf{x} \in \mathbb{R}^d$ denotes the multi-source remote sensing input data, $\hat{y} \in \mathbb{R}$ denotes the estimated physical parameter, $f_\theta$ is a learnable nonlinear mapping function, and $\mathcal{L}$ is the loss function quantifying the inversion error. To approximate this complex nonlinear relationship, inversion models often leverage radiative transfer models, look-up tables (LUT), machine learning regressors, or deep neural networks \cite{qianqian2022research, li2023bessv2}.

From the input perspective, quantitative inversion relies on data acquired by multimodal sensors such as optical, multispectral, hyperspectral, thermal infrared, synthetic aperture radar (SAR), and LiDAR. These sensors differ significantly in spectral range, spatial resolution, viewing geometry, and revisit frequency, creating a highly heterogeneous and high-dimensional input space. From the output perspective, the target variables typically exhibit spatial continuity, temporal dynamics, and cross-regional transferability. Therefore, models are expected to produce not only high-accuracy pointwise estimations but also spatially consistent and physically plausible predictions across large-scale geographical regions \cite{zhu2017deep, bai2024integrating}.

With the continuous advancement in remote sensing acquisition capabilities and the rapid evolution of artificial intelligence, quantitative inversion tasks are transitioning from traditional small-scale, single-variable modeling to multi-variable, multi-task, and cross-modal collaborative modeling. This paradigm shift demands more sophisticated model architectures and has prompted researchers to pay increased attention to scientific consistency, generalization capacity, and the integration of data-driven and physics-based approaches. These developments are collectively propelling remote sensing inversion from the perception layer to the cognition layer \cite{li2023big, qianqian2022research, nedungadi2024mmearth}.

\subsection{Task Categories and Representative Variables}

Quantitative remote sensing inversion involves the continuous estimation of physically meaningful surface variables that describe key ecological, hydrological, and atmospheric processes. These variables serve as essential inputs for Earth system modeling, environmental assessment, and resource management. Based on the underlying geophysical processes and domain-specific applications, inversion tasks can be broadly categorized into five representative groups: vegetation structural parameters, biomass and carbon estimation, hydrological and energy fluxes, surface physical properties, and water quality and atmospheric composition. Table~\ref{tab:landscape_variables} summarizes these categories along with representative variables, full names, and typical application scenarios.

\textbf{Vegetation structural parameters} such as LAI, NDVI, FAPAR, and FVC characterize canopy density, photosynthetic capacity, and vegetation cover. These parameters are foundational for monitoring vegetation dynamics, crop yield prediction, and ecosystem modeling. They are often estimated using canopy radiative transfer models such as PROSAIL and SCOPE, which simulate spectral responses across varying canopy structures.

\textbf{Biomass and carbon estimation} tasks target parameters such as aboveground biomass (AGB), canopy height (CH), and carbon stock (CS). These variables are key to forest carbon accounting, biomass inventorying, and climate change policy evaluation. Models such as DART and FLIGHT simulate 3D canopy structure and light scattering, while allometric and empirical models are used to relate remote sensing observations to biomass and carbon content.

\textbf{Hydrological and energy flux variables}, including evapotranspiration (ET), soil moisture (SM), land surface temperature (LST), and vapor pressure deficit (VPD), reflect surface water-energy exchange dynamics. These parameters are crucial for hydrological modeling, drought monitoring, and climate response assessments. Energy balance models such as SCOPE and Penman–Monteith, as well as land surface models (e.g., Noah, CLM), support the inversion of these variables based on thermal or microwave observations.

\textbf{Surface physical properties} such as albedo, surface roughness, and land surface emissivity (LSE) influence radiative transfer and energy balance at the land–atmosphere interface. Accurate retrieval of these properties enables improved surface parameterization in climate and energy models. Their estimation typically involves models like PROSAIL, DART, and MODTRAN.

\textbf{Water quality and atmospheric composition} variables—including turbidity, total suspended solids (TSS), colored dissolved organic matter (CDOM), and aerosol optical depth (AOD)—are indicative of aquatic ecosystem health and atmospheric clarity. Hydrolight and bio-optical models simulate water-leaving radiance under varying water quality conditions, while MODTRAN and 6S serve as standard atmospheric radiative transfer models for aerosol characterization.

These inversion tasks are typically defined over continuous variable domains, exhibit multi-scale spatial and temporal dynamics, and are derived from multimodal remote sensing data. Moreover, they are often coupled—e.g., LAI and FAPAR, AGB and CH, or ET and SM—reflecting interdependent biophysical processes. %As shown in Table~\ref{tab:landscape_variables}, these tasks span diverse application scenarios and impose significant challenges for model design, including generalization across regions, sensor modalities, and environmental conditions.

% % 表 1. 遥感定量反演任务分类与代表性变量

\begin{landscape}
\begin{table}[!ht]
\centering
\caption{Remote Sensing Inversion Tasks with Categories, Variables, Full Names, and Applications}
\label{tab:landscape_variables}
\begin{tabular}{c|p{3cm}|p{1.5cm}|p{5.5cm}|p{6.5cm}}
\toprule
\textbf{No.} & \textbf{Category} & \textbf{Variable} & \textbf{Full Name} & \textbf{Application Scenarios} \\
\midrule

\multirow{4}{*}{1} & \multirow{4}{=}{Vegetation Structural Parameters} 
& LAI & Leaf Area Index & \multirow{4}{=}{Vegetation dynamics monitoring, ecosystem modeling, drought detection, crop yield estimation} \\
& & NDVI & Normalized Difference Vegetation Index & \\
& & FAPAR & Fraction of Absorbed Photosynthetically Active Radiation & \\
& & FVC & Fractional Vegetation Cover & \\

\midrule

\multirow{4}{*}{2} & \multirow{4}{=}{Biomass and Carbon Estimation} 
& AGB & Aboveground Biomass & \multirow{4}{=}{Forest carbon stock assessment, biomass inventory, carbon neutrality and policy modeling} \\
& & CH & Canopy Height & \\
& & CS & Carbon Stock & \\
& & Biomass & Biomass Index & \\

\midrule

\multirow{4}{*}{3} & \multirow{4}{=}{Hydrological and Energy Fluxes} 
& ET & Evapotranspiration & \multirow{4}{=}{Hydrological cycle simulation, flood management, climate impact analysis} \\
& & SM & Soil Moisture & \\
& & LST & Land Surface Temperature & \\
& & VPD & Vapor Pressure Deficit & \\

\midrule

\multirow{3}{*}{4} & \multirow{3}{=}{Surface Physical Properties} 
& Albedo & Surface Reflectance Coefficient & \multirow{3}{=}{Radiative transfer modeling, energy balance modeling, land cover classification} \\
& & Roughness & Surface Roughness & \\
& & LSE & Land Surface Emissivity & \\

\midrule

\multirow{4}{*}{5} & \multirow{4}{=}{Water Quality and Atmospheric Composition} 
& Turbidity & Water Turbidity & \multirow{4}{=}{Water quality monitoring, eutrophication assessment, pollutant tracing, atmospheric correction} \\
& & TSS & Total Suspended Solids & \\
& & CDOM & Colored Dissolved Organic Matter & \\
& & AOD & Aerosol Optical Depth & \\

\bottomrule
\end{tabular}
\end{table}
\end{landscape}

\subsection{Key Characteristics}

Quantitative remote sensing inversion exhibits distinct methodological and structural characteristics that differentiate it fundamentally from conventional remote sensing tasks such as classification or object detection. As summarized in Table~\ref{tab:task_properties}, its core attributes span five major dimensions: output continuity, spatial and temporal scalability, multimodal input structures, and cross-regional generalizability.

\textbf{(1) Output continuity and physical interpretability.}  
In contrast to categorical prediction tasks, quantitative inversion aims to estimate continuous-valued geophysical variables (e.g., LAI, soil moisture, evapotranspiration) that possess explicit physical meanings and ecological relevance \cite{li2023big}. These variables are intrinsically linked to underlying surface processes, and thus the inversion models must not only achieve high numerical accuracy but also ensure physical consistency and semantic interpretability \cite{qianqian2022research}. This imposes additional constraints on model formulation, loss design, and evaluation metrics.

\textbf{(2) Multi-scale spatial and temporal representation.}  
Inversion tasks require models to operate across a broad range of spatial scales—from high-resolution optical sensors (e.g., Sentinel-2 at 10 m) to medium- and low-resolution platforms (e.g., MODIS at 500 m)—necessitating effective scale normalization and resolution-aware modeling strategies \cite{zhu2017deep}. Temporally, inversion must support both snapshot-based estimation and dynamic modeling over seasonal or interannual timescales, requiring robust temporal encoding and sequential representation learning capabilities \cite{han2023survey}.

\textbf{(3) Multimodal data fusion and sensor heterogeneity.}  
Modern Earth observation platforms generate multimodal datasets across diverse sensor types, including multispectral (MSI), hyperspectral (HSI), synthetic aperture radar (SAR), thermal infrared (TIR), and LiDAR. These modalities offer complementary information in terms of spectral sensitivity, structural penetration, and material discrimination \cite{bai2024integrating}. For instance, SAR provides structural features under all-weather conditions, while HSI enables fine-grained material classification. Effectively integrating such heterogeneous inputs remains a critical challenge for inversion models \cite{nedungadi2024mmearth}.

\textbf{(4) Cross-region generalization under domain shifts.}  
Remote sensing inversion often involves geographic domain shifts due to variability in topography, vegetation type, and climatic conditions between training and application regions \cite{peng2022domain}. This leads to degradation in model performance when deployed beyond the training distribution. Achieving cross-region transferability therefore requires models to be robust to distributional discrepancies, which motivates the adoption of weak supervision, domain adaptation, multi-task learning, and foundation model pretraining strategies \cite{hu2025dualstrip, nedungadi2024mmearth}.

\textbf{(5) Integrated modeling constraints and transfer mechanisms.}  
Given the inherent heterogeneity in variable types, spatial resolution, and sensor modalities, inversion models must satisfy multiple constraints simultaneously: accurate numerical regression, physical plausibility, spatiotemporal consistency, and transferability. This necessitates hybrid modeling strategies that integrate physical priors, representation learning, and knowledge-guided regularization into a unified framework for robust and scalable inversion.

\begin{table}[ht]
\centering
\caption{Core Characteristics and Challenges of Quantitative Remote Sensing Inversion}
\label{tab:task_properties}
\begin{tabular}{p{3.5cm}|p{4cm}|p{5.5cm}}
\toprule
\textbf{Characteristic} & \textbf{Representation} & \textbf{Challenges} \\
\midrule
Output Variable Type & Continuous values with physical semantics & 
1. Achieving accurate regression \newline
2. Preserving physical consistency \\ \midrule

Spatial Resolution & Multi-scale (10--500 m) & 
1. Handling scale variation \newline
2. Designing resolution-adaptive models \\ \midrule

Temporal Dynamics & Single to multi-temporal sequences & 
1. Modeling seasonal dynamics \newline
2. Learning temporal dependencies \\ \midrule

Data Modalities & Multisource and heterogeneous sensors & 
1. Sensor alignment and fusion \newline
2. Capturing modality-specific structures \\ \midrule

Geographic Generalization & Cross-region deployment & 
1. Mitigating domain shifts \newline
2. Ensuring transferability and robustness \\ 
\bottomrule
\end{tabular}
\end{table}

\section{Physics-Based Methods}
\subsection{Theoretical and Mechanistic Foundations}

Physics-based methods in quantitative remote sensing inversion are grounded in the simulation of fundamental physical processes governing the interaction of electromagnetic radiation with the Earth's surface and atmosphere. These approaches aim to preserve physical consistency and semantic interpretability by explicitly modeling radiative transfer, surface energy balance, biochemical dynamics, or hydrodynamic behaviors, depending on the variable of interest.

Among these, Radiative Transfer Models (RTMs) remain central to characterizing radiation–vegetation interactions. The PROSAIL model, which couples the PROSPECT module (leaf optical properties) with SAIL (canopy scattering), enables the forward simulation of directional reflectance spectra and supports inversion of vegetation structural parameters such as LAI, NDVI, FAPAR, and FVC \cite{jacquemoud2009prospect, verrelst2015optical}. For more complex canopies and heterogeneous surfaces, the DART model simulates three-dimensional radiative transfer and has been extensively applied to the estimation of aboveground biomass (AGB) and canopy height (CH) \cite{gastellu2004dart}.

SCOPE (Soil Canopy Observation, Photochemistry and Energy fluxes) extends this framework by incorporating photosynthesis, thermal radiation, and energy balance processes, thereby facilitating the joint inversion of variables such as evapotranspiration (ET), land surface temperature (LST), solar-induced fluorescence (SIF), and vapor pressure deficit (VPD) \cite{yang2021scope}. In aquatic and atmospheric domains, models such as Hydrolight simulate underwater light propagation in relation to turbidity, total suspended solids (TSS), and colored dissolved organic matter (CDOM) \cite{zaneveld1995light, binding2008spectral}. Atmospheric radiative transfer tools including MODTRAN and 6S are employed for aerosol optical depth (AOD) estimation and atmospheric correction by modeling radiation paths through different atmospheric layers \cite{vermote1997second, verhoef2003simulation}.

In addition to RTMs, various mechanistic models are used for specific geophysical targets. For instance, allometric models, which relate tree structure to biomass, are commonly adopted in forest carbon estimation \cite{li2022toward}. The Penman–Monteith equation is widely used in ET modeling, while soil moisture retrieval is often based on dielectric mixing models or land surface schemes such as Noah and CLM \cite{li2023big}.

Table~\ref{tab:physics_variable_mapping} summarizes key remote sensing inversion variables alongside their representative mechanistic models. These frameworks collectively constitute the theoretical foundation of physics-based inversion, offering interpretability, transferability, and robustness, particularly under data-limited or cross-regional conditions.

% 表2. 常见遥感反演的机理模型概览
\begin{table}[ht]
\centering
\caption{Mapping of Remote Sensing Inversion Variables to Representative Mechanistic Models}
\label{tab:physics_variable_mapping}
\small
\begin{tabular}{c|c|c}
\toprule
\textbf{Variable} & \textbf{Representative Mechanistic Models} & \textbf{References} \\
\midrule

\multicolumn{3}{l}{\textit{1. Vegetation Structural Parameters}} \\
\midrule
LAI & PROSAIL, SCOPE, DART & \cite{jacquemoud2009prospect, yang2021scope, gastellu2004dart} \\
NDVI & PROSAIL (reflectance proxy) & \cite{jacquemoud2009prospect} \\
FAPAR & SAIL, SCOPE & \cite{verrelst2015optical, yang2021scope} \\
FVC & SAIL, PROSAIL & \cite{verrelst2015optical} \\

\midrule
\multicolumn{3}{l}{\textit{2. Biomass and Carbon Estimation}} \\
\midrule
AGB & DART, FLIGHT, allometric biomass models & \cite{gastellu2004dart, north2002three, li2022toward} \\
CH & DART, LiDAR-based canopy models & \cite{north2002three} \\
CS & AGB × carbon coefficient models (e.g., IPCC) & \cite{guo2021mapping} \\
Biomass & Spectral-structural index models & \cite{li2022toward} \\

\midrule
\multicolumn{3}{l}{\textit{3. Hydrological and Energy Fluxes}} \\
\midrule
ET & SCOPE, Penman–Monteith equation & \cite{yang2021scope} \\
LST & SCOPE (thermal balance), surface radiative models & \cite{yang2021scope} \\
SM & Microwave dielectric models, land surface models (Noah, CLM) & \cite{li2023big} \\
VPD & Meteorological + canopy energy models (via SCOPE) & \cite{yang2021scope} \\

\midrule
\multicolumn{3}{l}{\textit{4. Surface Physical Properties}} \\
\midrule
Albedo & PROSAIL, SCOPE & \cite{jacquemoud2009prospect, yang2021scope} \\
Roughness & DART, Li–Strahler geometric–optical model & \cite{gastellu2004dart} \\
LSE & SCOPE, MODTRAN (thermal emissivity) & \cite{yang2021scope, vermote1997second} \\

\midrule
\multicolumn{3}{l}{\textit{5. Water Quality and Atmospheric Composition}} \\
\midrule
Turbidity & Hydrolight, semi-analytical bio-optical models & \cite{zaneveld1995light, bernardo2019retrieval} \\
TSS & Hydrolight, empirical scattering-based models & \cite{bernardo2019retrieval} \\
CDOM & Hydrolight, absorption-based bio-optical models & \cite{binding2008spectral} \\
AOD & 6S, MODTRAN (atmospheric radiative transfer) & \cite{verhoef2003simulation, vermote1997second} \\

\bottomrule
\end{tabular}
\end{table}

% \subsection{Parameter Inversion Pathways}
\subsection{Inverse Modeling Strategies}

Physics-based inversion in remote sensing commonly adopts a forward–inverse modeling paradigm, wherein the forward model simulates the transformation from surface biophysical parameters to observable signals, such as spectral reflectance or radiance. Inversion strategies then aim to retrieve the unknown parameters by inverting this model. Depending on how the inversion process is formulated and solved, three principal categories of approaches can be distinguished.

\textbf{Look-Up Table (LUT) Methods} construct a synthetic database of reflectance or radiance values across the parameter space using forward simulations (e.g., RTMs). During inversion, the observed signal is compared against this table to identify the best-matching entry. LUT methods are easy to implement, do not require model differentiability, and allow for fast prototyping \cite{verrelst2015optical}. Nevertheless, their applicability is constrained by sampling resolution, high computational cost in high-dimensional settings, and poor generalization to out-of-distribution scenarios.

\textbf{Optimization-Based Estimation} reformulates inversion as a minimization problem over a defined cost or likelihood function, such as least-squares error or Bayesian posterior. Estimation is then performed using iterative algorithms, including gradient descent, Markov Chain Monte Carlo (MCMC), or Kalman filtering \cite{combal2003retrieval}. These methods offer high accuracy when applied to well-posed, differentiable models. However, they are often computationally demanding and susceptible to local minima and sensitivity to initialization, limiting their scalability in large or noisy datasets.

\textbf{Physics-Informed Neural Networks (PINNs)} represent a more recent class of methods that integrate physical equations or constraints—such as radiative transfer or energy balance—into the training of deep neural networks. Rather than replacing physical models, PINNs embed them as soft constraints or loss terms, thereby improving learning efficiency and physical plausibility \cite{karpatne2017theory}. These models are particularly suited for weakly supervised or data-scarce applications. Their limitations include instability during training, sensitivity to loss term weighting, and challenges in interpretability.

Taken together, these strategies illustrate the evolution of physics-based inversion from table-driven and iterative methods toward integrated frameworks that leverage both physical knowledge and statistical learning. The comparative characteristics of the three approaches are summarized in Table~\ref{tab:physics_methods_comparison}.

\begin{table}[ht]
\centering
\caption{Comparison of Physics-Based Inversion Strategies}
\label{tab:physics_methods_comparison}
\begin{tabular}{p{2.5cm}|p{4.5cm}|p{4.5cm}}
\toprule
\textbf{Strategy} & \textbf{Advantages} & \textbf{Limitations} \\
\midrule

\textbf{Look-Up Table (LUT)} \cite{verrelst2015optical} &
\begin{itemize}
  \item Easy to implement and interpret
  \item No gradient required; applicable to non-differentiable models
  \item Minimal reliance on model assumptions
\end{itemize} &
\begin{itemize}
  \item Limited resolution and scalability
  \item Poor generalization to out-of-distribution data
  \item High computational cost in high-dimensional settings
\end{itemize} \\

\midrule

\textbf{Optimization-Based Estimation} \cite{combal2003retrieval} &
\begin{itemize}
  \item High retrieval accuracy under well-posed conditions
  \item Compatible with probabilistic and Bayesian frameworks
  \item Flexible for differentiable physical models
\end{itemize} &
\begin{itemize}
  \item Sensitive to initial values and local optima
  \item Computationally demanding for large-scale problems
  \item Limited robustness in ill-posed or noisy scenarios
\end{itemize} \\

\midrule

\textbf{Physics-Informed Neural Networks (PINNs)} \cite{karpatne2017theory} &
\begin{itemize}
  \item Embeds physical laws into learning
  \item Strong expressive power with data-physics synergy
  \item Effective in low-sample or weak-supervision settings
\end{itemize} &
\begin{itemize}
  \item Requires careful loss term balancing
  \item May suffer from convergence instability
  \item Interpretability and debugging remain challenging
\end{itemize} \\

\bottomrule
\end{tabular}
\end{table}

\subsection{Challenges}

\textbf{(1) Limited observability of key physical parameters.}
Physics-based inversion methods offer strong physical grounding and interpretability, as they directly model the causal relationships between surface states and observed signals \cite{hu2023physical}. However, their reliance on parameters that are not directly measurable—such as leaf biochemical constituents, canopy architecture, or sub-surface soil properties—often necessitates empirical assumptions or indirect proxies. This limits model robustness and introduces uncertainty, particularly in heterogeneous or poorly characterized environments.

\textbf{(2) High computational burden restricts operational scalability.}
Simulations based on RTMs like PROSAIL or SCOPE typically involve solving nonlinear, high-dimensional equations with radiative or thermodynamic constraints. While such models are theoretically complete, they are computationally intensive and often unsuitable for large-scale or real-time inversion. Recent efforts to accelerate RTM simulations using emulators or surrogate models remain limited in generalizability \cite{brodrick2021generalized}. The associated computational cost is a key barrier to integrating these models into Earth system applications that demand efficiency and throughput.

\textbf{(3) Simplifying assumptions limit model generalization.}
Many RTMs are built upon idealized assumptions—such as horizontally homogeneous canopies, isotropic scattering, or stable atmospheric layers—which may hold in experimental plots but not in real-world landscapes. These simplifications compromise model performance in complex terrains, mixed vegetation types, or anisotropic illumination conditions, resulting in limited geographic transferability \cite{bai2024integrating}.

\textbf{(4) Poor adaptability in weak-supervision and multi-source settings.}
Purely physical models typically assume high-quality, domain-specific input and calibration. In contrast, many practical remote sensing applications operate under weak supervision, limited ground truth, and multi-sensor fusion (e.g., combining optical, thermal, and SAR data). In such contexts, RTMs lack the flexibility to adapt across modalities and conditions without extensive reparameterization or prior knowledge \cite{li2023big}.

\textbf{(5) Data-driven integration remains challenging but necessary.}
To overcome the limitations above, integrating physical models with machine learning is increasingly seen as a promising direction \cite{han2023survey}. Nonetheless, this hybridization poses new challenges, such as how to encode physical constraints into differentiable architectures, or how to balance physical priors with data-driven patterns in the loss landscape. This remains an open question in current research, especially in global-scale or ill-posed inversion settings \cite{karpatne2017theory, han2023survey}.

\section{Data-Driven Methods}
\subsection{Machine Learning Methods}

The development of traditional machine learning (ML) algorithms has provided a pragmatic solution for remote sensing inversion, particularly in scenarios where physical models face limitations in scalability, parameter observability, or computational efficiency. With the increasing availability of labeled geospatial data, empirical learning frameworks such as Random Forests (RF), Support Vector Regression (SVR), and gradient boosting techniques (e.g., XGBoost) have been widely adopted to estimate land surface variables from multi-source remote sensing inputs \cite{belgiu2016random, han2023survey}.

RF models are particularly favored for their robustness, low sensitivity to parameter tuning, and inherent feature importance measures. They have been extensively used for variables including leaf area index (LAI), fraction of absorbed photosynthetically active radiation (FAPAR), and aboveground biomass (AGB), often with satisfactory accuracy under varying ecological conditions \cite{soltanikazemi2022field}. SVR, by contrast, offers high generalization performance in small-sample and high-dimensional settings, making it suitable for localized or region-specific inversion tasks where data scarcity remains a challenge \cite{han2023survey}. XGBoost and other ensemble-based boosting methods have demonstrated improved prediction stability and nonlinear regression capacity, particularly in the estimation of land surface temperature (LST), soil properties, and aquatic parameters \cite{li2015estimating, li2024machine}.

Despite these advantages, traditional ML methods remain constrained by their reliance on manually derived features and statistical descriptors, which may not capture the full complexity of raw image data. Their performance often degrades when applied to high-dimensional inputs such as hyperspectral imagery or when spatial–temporal dynamics are prominent. Moreover, conventional ML models typically lack the architectural capacity to integrate heterogeneous inputs (e.g., SAR and optical data), limiting their effectiveness in multimodal or multi-scale data fusion tasks.

Recent studies have explored hybrid strategies, such as feature-enhanced regression using texture indices or dimensionality-reduction preprocessing (e.g., PCA, t-SNE), to partially alleviate these constraints. However, the absence of end-to-end representation learning remains a core limitation. As highlighted in Table~\ref{tab:ml_dl_fm_comparison} and Table~\ref{tab:rs_inversion_models_final}, ML models continue to play a complementary role in modern inversion pipelines, especially when interpretability, lightweight deployment, or limited training samples are prioritized.

\subsection{Deep Learning Methods}

The rise of deep learning has profoundly transformed remote sensing inversion, particularly in applications requiring spatial, temporal, or spectral feature extraction from high-dimensional imagery. Deep neural networks (DNNs), especially those with convolutional or recurrent structures, have demonstrated superior capacity in modeling nonlinear and multi-modal relationships that traditional approaches struggle to capture.

Convolutional Neural Networks (CNNs) are among the most widely used architectures for pixel- and patch-level inversion of biophysical parameters such as NDVI, ET, and AGB. Their convolutional structure enables spatially consistent learning from optical or multispectral imagery, making them particularly effective in vegetation monitoring and land surface characterization \cite{zhu2017deep, su2024deep}. In contrast, Recurrent Neural Networks (RNNs) and their improved variant, Long Short-Term Memory (LSTM) networks, are well-suited for modeling sequential dependencies in time-series datasets. These models have been successfully applied to MODIS or Sentinel-based multi-temporal datasets for tasks such as evapotranspiration prediction and drought monitoring \cite{ma2022forecasting, zheng2024predicting}.

Recent advances have also introduced Graph Neural Networks (GNNs) for learning spatial dependencies beyond fixed grids, particularly useful in applications involving irregular geographic entities or network-structured inputs. Meanwhile, Transformers, originally designed for sequence modeling, have shown notable effectiveness in long-range spatial–temporal learning. Their self-attention mechanism enables global contextual understanding and has recently been adapted for remote sensing inversion, including canopy structure estimation and carbon stock modeling \cite{dosovitskiy2020image, rehman2025deep}.

Another key strength of deep learning lies in its support for end-to-end training pipelines. This enables joint optimization from raw input to prediction, reducing the need for handcrafted features. Furthermore, strategies such as transfer learning, self-supervised pretraining, and generative data augmentation (e.g., GANs) are increasingly used to mitigate data scarcity and improve robustness in under-sampled regions \cite{han2023survey}. CNN–LSTM hybrid models and attention-based multi-branch architectures also facilitate the integration of heterogeneous inputs such as SAR, optical, and meteorological data \cite{xie2022combining}.

Nonetheless, deep models are not without limitations. Their lack of interpretability, high demand for labeled training data, and sensitivity to regional domain shifts remain persistent challenges. While visualization tools and explainability techniques (e.g., saliency maps, SHAP) are under active development, their maturity in geoscientific applications remains limited. Moreover, generalization performance often deteriorates when models are transferred across regions with differing land cover, sensor conditions, or phenological cycles, limiting their scalability for global applications.

In sum, deep learning has enabled substantial improvements in estimation accuracy and model capacity, yet its reliability, transparency, and regional transferability remain open concerns. The integration of deep models with domain knowledge and physically informed constraints represents a promising avenue for future development.

\subsection{Foundation Models}

Foundation models (FMs) have recently emerged as a transformative paradigm in quantitative remote sensing inversion, characterized by large-scale pretraining, multi-modal integration, and task-agnostic adaptation. Unlike conventional deep learning approaches that rely on task-specific architectures and extensive supervision, FMs are designed to learn unified spatial–temporal representations across variables, tasks, and sensor modalities. These models typically adopt masked modeling or contrastive objectives during pretraining, enabling them to capture generic geospatial priors and support a variety of downstream inversion tasks under limited supervision.

Representative models include \textbf{SatMAE}, which employs masked autoencoding over multi-temporal Sentinel-2 data to learn spectral–temporal representations. Its encoder–decoder structure facilitates fine-tuning for variables such as NDVI, LAI, and FAPAR, demonstrating improved performance in label-scarce settings~\cite{cong2022satmae}. \textbf{GFM (Geospatial Foundation Model)} builds on this idea by introducing a multimodal Transformer backbone capable of fusing optical, SAR, and DEM data. Through a shared latent representation and multi-task adaptation head, GFM supports concurrent estimation of AGB, CH, SIF, and CS, and exhibits strong generalization across ecological regions~\cite{mendieta2023towards}. Extending further, \textbf{mmEarth} aligns diverse modalities including hyperspectral imagery, SAR, and DSM using contrastive and multimodal masking strategies. Its architecture is optimized for pixel-level regression under heterogeneous input spaces, addressing domain shifts and multi-source inconsistencies~\cite{nedungadi2024mmearth}.

Despite these advances, current foundation models face several limitations in quantitative inversion contexts. First, the design of regression heads remains underexplored, with most approaches relying on shallow MLPs that may not capture the task-specific structure of geophysical variables. Second, fine-tuning strategies often require substantial computational resources and lack standardization, making adaptation across tasks or regions non-trivial. Finally, most existing FMs treat remote sensing inversion as a purely data-driven problem, with minimal incorporation of physical constraints, leading to concerns regarding physical plausibility and interpretability. Ongoing research is exploring physically-informed pretraining, adaptive regression heads, and uncertainty-aware modeling to bridge this gap and build more trustworthy inversion systems.

\begin{table}[ht]
\centering
\caption{Overview of Foundation Models for Quantitative Remote Sensing Inversion}
\label{tab:foundation_models}
\begin{tabular}{p{1.2cm}|p{2cm}|p{2.8cm}|p{3.8cm}|p{3cm}}
\toprule
\textbf{Model} & \textbf{Backbone} & \textbf{Input Modalities} & \textbf{Pretraining Strategy} & \textbf{Supported Tasks} \\
\midrule
SatMAE \cite{cong2022satmae} & MAE + Vision Transformer & Multi-temporal Sentinel-2 & Masked Autoencoding (masking + reconstruction) & NDVI, LAI, FAPAR  \\
\midrule
GFM \cite{mendieta2023towards} & Multimodal Transformer & Optical + SAR + DEM & Self-supervised + Multi-task learning & AGB, CH, SIF, CS \\
\midrule
mmEarth \cite{nedungadi2024mmearth} & Cross-modal Transformer & HSI + SAR + DSM & Masked + Contrastive Learning & NDVI, Biomass \\
\bottomrule
\end{tabular}
\end{table}

\begin{table}[ht]
\centering
\caption{Comparison of Machine Learning, Deep Learning, and Foundation Models for Remote Sensing Inversion}
\label{tab:ml_dl_fm_comparison}
\begin{tabular}{p{3cm}|p{4.5cm}|p{4.5cm}}
\toprule
\textbf{Method Category} & \textbf{Strengths} & \textbf{Limitations} \\
\midrule

\textbf{Machine Learning} \newline (e.g., RF, SVR, XGBoost) &
\begin{itemize}
  \item Interpretable and easy to implement
  \item Performs well with small or moderate datasets
  \item Effective for structured, tabular-like inputs
\end{itemize} &
\begin{itemize}
  \item Requires manual feature extraction
  \item Poor in handling high-dimensional raw imagery
  \item Weak in capturing spatial or temporal context
\end{itemize} \\

\midrule

\textbf{Deep Learning} \newline (e.g., CNN, LSTM, Transformer) &
\begin{itemize}
  \item Learns directly from raw spatial/temporal data
  \item Captures nonlinear and multimodal relationships
  \item Supports transfer learning and data augmentation
\end{itemize} &
\begin{itemize}
  \item Demands large labeled datasets
  \item Limited interpretability in scientific contexts
  \item Generalization across regions remains challenging
\end{itemize} \\

\midrule

\textbf{Foundation Models} \newline (e.g., SatMAE, GFM, mmEarth) &
\begin{itemize}
  \item Pretrained on large-scale multi-source data
  \item Strong capability in multi-task and multi-modal fusion
  \item Suited for weak supervision and few-shot adaptation
\end{itemize} &
\begin{itemize}
  \item High computational cost for training and inference
  \item Regression adaptation not always straightforward
  \item Fine-tuning strategies still under development
\end{itemize} \\

\bottomrule
\end{tabular}
\end{table}

\begin{landscape}
\begin{table}[ht]
\centering
\caption{Mapping of Inversion Variables to ML, DL, and FM Models with Input Data}
\label{tab:rs_inversion_models_final}
\small
\begin{tabular}{c|c|c|c|c|c}
\toprule
\textbf{Category} & \textbf{Variable} & \textbf{Input Data} & \textbf{ML Models} & \textbf{DL Models} & \textbf{FM Models} \\
\midrule

\multirow{4}{*}{Vegetation Structure}
& LAI   & Sentinel‑2, SAR         & RF \cite{soltanikazemi2022field}, B‑RF \cite{zhang2024retracted}    & CNN–LSTM \cite{xie2022combining}        & SatMAE \cite{cong2022satmae}         \\
& NDVI  & Sentinel‑2              & RF \cite{soltanikazemi2022field}                              & CNN \cite{su2024deep}         & mmEarth \cite{nedungadi2024mmearth}    \\
& FAPAR & MODIS, Sentinel‑2       & RF \cite{han2023survey}                              & ResNet \cite{xie2022combining}         & SatMAE \cite{cong2022satmae}         \\
& FVC   & Landsat, Vegetation inds.& SVR \cite{han2023survey}                             & CNN \cite{zhu2017deep}          & —                                     \\
\midrule

\multirow{4}{*}{Biomass \& Carbon}
& AGB     & Sentinel, Lidar         & XGBoost \cite{li2024machine}                                & AGBUNet \cite{arumai2025agbunet}   & GFM \cite{mendieta2023towards}           \\
& CH      & Lidar, RGB              & SVR \cite{li2022toward}                       & CNN \cite{cao2024deep}        & GFM \cite{mendieta2023towards}           \\
& CS      & AGB × Coef.             & RF \cite{guo2021mapping}                           & GBDT \cite{li2022toward}   & GFM \cite{mendieta2023towards}           \\
& Biomass & Sentinel, Canopy metrics& RF \cite{li2024machine}                                & CNN \cite{zhu2017deep}          & mmEarth \cite{nedungadi2024mmearth}    \\
\midrule

\multirow{4}{*}{Hydrology \& Energy}
& ET   & MODIS, meteo             & RF \cite{di2024estimation}                                & LSTM \cite{zheng2024predicting}          & —                                     \\
& LST  & MODIS, Landsat           & XGBoost \cite{li2015estimating}                        & CNN–LSTM \cite{han2023survey}     & mmEarth \cite{nedungadi2024mmearth}    \\
& SM   & Microwave, NDVI          & SVR \cite{rains2021sentinel}                              & LSTM \cite{efremova2021soil}           & —                                     \\
& VPD  & S2, Meteorology          & RF \cite{chen2024prediction}                               & CNN \cite{xie2022combining}             & —                                     \\
\midrule

\multirow{3}{*}{Surface Properties}
& Albedo    & MODIS, BRDF comps      & RF \cite{li2015estimating}                         & CNN \cite{zhu2017deep}           & —                                     \\
& Roughness & DEM, Lidar-derived     & GBDT \cite{fan2023using}                            & CNN \cite{zheng2024predicting}           & —                                     \\
& LSE       & TIR, NDVI              & RF \cite{zhao2019practical}                               & CNN \cite{su2024deep}          & —                                     \\
\midrule

\multirow{4}{*}{Water \& Atmosphere}
& Turbidity & HSI, MSI               & RF \cite{adjovu2023measurement}                                & CNN \cite{moon2024deep}            & —                                     \\
& TSS       & Sentinel‑2, Rivers id. & SVR \cite{bernardo2019retrieval}                   & CNN \cite{moon2024deep}            & —                                     \\
& CDOM      & Optical reflectance    & XGBoost \cite{harkort2023estimation}                            & CNN \cite{harkort2023estimation}          & —                                     \\
& AOD       & MODIS, meteorology      & RF \cite{soltanikazemi2022field}                              & CNN \cite{zheng2024predicting}           & mmEarth \cite{nedungadi2024mmearth}    \\
\bottomrule
\end{tabular}
\end{table}
\end{landscape}

\subsection{Challenges}

Despite the substantial progress brought by data-driven methods—especially the emergence of foundation models—in improving accuracy and flexibility for remote sensing inversion, several challenges remain that hinder their scientific reliability and large-scale deployment:

\textbf{1. Limited interpretability and physical consistency.}  
Most machine learning and deep learning models rely on statistical associations rather than physically grounded mechanisms. The absence of domain constraints and causal structures makes them difficult to interpret or validate in geophysical applications, raising concerns about their scientific credibility and error propagation \cite{li2023big, qianqian2022research}.

\textbf{2. Dependency on high-quality labeled samples.}  
Inversion tasks such as AGB, ET, and SM estimation require accurate field measurements for supervision, yet such ground truth is often scarce, expensive to acquire, and spatially biased. While pretraining and transfer learning can mitigate this to some extent, model performance still heavily depends on sample density and representativeness across regions and scales \cite{di2024estimation, han2023survey}.

\textbf{3. Domain shift and poor generalization.}  
Data-driven models often fail to maintain accuracy under cross-regional, cross-seasonal, or sensor-inconsistent scenarios. This generalization bottleneck—especially pronounced in ecological or climate-sensitive variables—limits their applicability in global-scale or long-term monitoring settings \cite{bai2024integrating, hu2025dualstrip}.

\textbf{4. Incomplete integration of physical models.}  
Although recent foundation models have improved cross-modal representation, most architectures still lack explicit incorporation of radiative transfer models, energy balance equations, or physical constraints. As a result, predictions may violate physical laws or respond incorrectly to out-of-distribution inputs, limiting their use in scientific decision-making \cite{li2021estimation}.

\textbf{5. Computational cost and tuning complexity.}  
Advanced deep learning and foundation models typically require substantial computational resources for training and inference. Additionally, fine-tuning these models for regression tasks remains non-trivial, often involving architecture adaptation, head design, and loss function balancing \cite{nedungadi2024mmearth}.

In sum, future advancements in remote sensing inversion will require a transition from purely data-driven modeling to hybrid frameworks that combine physical consistency, interpretability, generalization, and efficiency. This direction holds the key to developing trustworthy, scalable, and explainable geospatial estimation systems.

\section{Datasets and Evaluation}
\textbf{High-quality datasets and scientifically sound evaluation metrics are fundamental to advancing quantitative remote sensing inversion methods.} In recent years, the emergence of open-access remote sensing data platforms and standardized inversion products has provided researchers with increasingly diverse, comprehensive, and physically consistent datasets. These resources enable the development of inversion models with enhanced comparability and generalizability. Concurrently, for inversion tasks involving continuous variable outputs, a performance evaluation framework has gradually taken shape. This framework encompasses multiple dimensions, including accuracy, structural preservation, and interpretability. This chapter systematically reviews the current technical foundations supporting mainstream inversion tasks, focusing on two key aspects: open data resources and evaluation metrics.

\subsection{Benchmark Datasets}

\textbf{Recent years have witnessed the emergence of open-access benchmark datasets tailored for remote sensing inversion, providing critical support for the development and evaluation of data-driven methods.} These datasets span a wide range of spatial resolutions (from 1 to 30 meters), geographic domains (regional to global), and sensor modalities (e.g., optical, SAR, LiDAR), enabling diverse estimation tasks across ecological, agricultural, and biogeophysical domains.

Table~\ref{tab:inversion_datasets} summarizes several representative datasets. \textbf{BigEarthNet} \cite{sumbul2021bigearthnet} is a large-scale benchmark combining Sentinel-1 and Sentinel-2 observations across Europe, widely used for land cover classification and vegetation index retrieval. \textbf{SSL4EO} \cite{wang2023ssl4eo} (including Sentinel-2 and Landsat variants) supports self-supervised learning and has been applied to LAI, FAPAR, and SIF estimation across multiple continents. \textbf{Satlas} \cite{bastani2023satlaspretrain} incorporates Sentinel-2, SMAP, and meteorological data for soil moisture and biomass inversion on a global scale, promoting fusion-based modeling. \textbf{Sen4AgriNet} \cite{sykas2021sen4agrinet} combines SAR and optical imagery with agronomic ground truth for crop and yield estimation in Europe and Africa. Finally, \textbf{Open-Canopy} \cite{fogel2025open} provides high-resolution RGB and LiDAR measurements, enabling detailed estimation of canopy height, aboveground biomass (AGB), and carbon stock at fine scales.

\textbf{Looking forward, the development of benchmark datasets with broader ecological coverage, richer annotation types, and stronger temporal depth will be essential} to support large-scale, multi-task, and foundation-model-based remote sensing inversion. In particular, datasets enabling the alignment of ground truth with physical priors and multi-modal observations will play a key role in advancing the accuracy, robustness, and interpretability of next-generation inversion models.

\begin{table}[ht]
\centering
\caption{Representative Datasets for Remote Sensing Inversion}
\label{tab:inversion_datasets}
\begin{tabular}{p{2cm}|p{1.5cm}|p{1cm}|p{3.5cm}|p{3.2cm}}
\toprule
\textbf{Dataset} & \textbf{Resolution} & \textbf{Region} & \textbf{Modalities} & \textbf{Inversion Tasks} \\
\midrule
\textbf{BigEarthNet} \cite{sumbul2021bigearthnet} & 10–20 m & Europe & Sentinel-1, Sentinel-2 & Land cover, NDVI, EVI \\
\midrule
\textbf{SSL4EO (S12/L)} \cite{wang2023ssl4eo,stewart2023ssl4eo} & 10–30 m & Global & Sentinel-1/2, Landsat & LAI, FAPAR, SIF \\
\midrule
\textbf{Satlas} \cite{bastani2023satlaspretrain} & 10–30 m & Global & Sentinel-2, SMAP, meteo & Soil moisture, AGB \\
\midrule
\textbf{Sen4AgriNet} \cite{sykas2021sen4agrinet} & 10–30 m & Europe, Africa & Sentinel-1/2 & Crops, Biomass \\
\midrule
\textbf{Open-Canopy} \cite{fogel2025open} & 1–10 m & France & RGB, LiDAR & CH, AGB, Carbon stock \\
\bottomrule
\end{tabular}
\end{table}

\subsection{Evaluation Metrics}

Remote sensing inversion tasks typically involve the prediction of continuous geophysical variables. Common evaluation metrics include Root Mean Square Error (RMSE) \cite{2014MomentsRMSE}, Mean Absolute Error (MAE) \cite{hodson2022rootMAE}, the Coefficient of Determination ($R^2$) \cite{chicco2021coefficientR2}, the Structural Similarity Index Measure (SSIM) \cite{mudeng2022prospectsSSIM}, and Peak Signal-to-Noise Ratio (PSNR) \cite{bae2024newPSNR}. RMSE and MAE respectively quantify the squared and absolute deviations between predictions and ground truth, and both converge to zero under perfect prediction, offering intuitive measures of average error.

The $R^2$ metric assesses the proportion of variance explained by the model and typically ranges from $(-\infty,\,1]$, with higher values indicating stronger explanatory power. Inversion tasks with spatially structured outputs—such as LAI or AGB maps—often require additional metrics to assess spatial consistency. SSIM evaluates similarity in terms of luminance, contrast, and structural information, where a score of 1 reflects perfect structural agreement. PSNR, expressed in decibels, is used to measure image fidelity; higher values suggest lower distortion and better detail preservation.
Together, these metrics provide a comprehensive evaluation framework covering error magnitude, variance explanation, and structural preservation, supporting fair comparison across different inversion models and tasks.

% 表 4. 常用遥感反演评价指标分类与说明
\begin{table}[ht]
\centering
\caption{Common Evaluation Metrics for Remote Sensing Inversion}
\label{tab:evaluation_metrics}
\begin{tabular}{c|l|c|c}
\toprule
\textbf{Abbreviation} & \textbf{Full Name} & \textbf{Typical Range} & \textbf{Optimal Value} \\
\midrule
RMSE \cite{2014MomentsRMSE} & Root Mean Square Error & $[0,\,+\infty)$ & 0 \\
MAE \cite{hodson2022rootMAE} & Mean Absolute Error & $[0,\,+\infty)$ & 0 \\
$R^2$ \cite{chicco2021coefficientR2} & Coefficient of Determination & $(-\infty,\,1]$ & 1 \\
SSIM \cite{mudeng2022prospectsSSIM} & Structural Similarity Index & $[0,\,1]$ & 1 \\
PSNR \cite{bae2024newPSNR} & Peak Signal-to-Noise Ratio (dB) & $[0,\,+\infty)$ & $\uparrow$ \\
\bottomrule
\end{tabular}
\end{table}

\section{Future Perspectives}

Although foundation model–driven data paradigms have substantially advanced the capabilities of quantitative remote sensing inversion, several core challenges remain unresolved. These challenges hinder the scientific credibility, operational scalability, and global adaptability of current inversion models. To move toward next-generation frameworks that are interpretable, generalizable, and robust across diverse scenarios, future research must address the following three priorities: interpretability and physical consistency, generalization and cross-domain adaptability, and data infrastructure and labeling quality.

\subsection{Interpretability and Physical Consistency}

\textbf{(1) Enhancing physical interpretability is crucial for scientific trust.} Most existing foundation models adopt end-to-end black-box architectures that directly map high-dimensional remote sensing data to target variables without incorporating underlying geophysical principles. This limits the interpretability and verifiability of model outputs, particularly in policy-sensitive or decision-critical applications \cite{karpatne2017theory, li2023big}. For example, predictions of carbon stock or evapotranspiration may lack physical plausibility if they do not adhere to known energy or mass conservation laws.

\textbf{(2) Future models should integrate physical constraints.} Embedding radiative transfer models, energy balance equations, or known functional relationships (e.g., LAI–NDVI, SM–SAR backscatter) within model architectures or training objectives can guide learning toward physically meaningful solutions. Approaches such as physics-informed loss functions, differentiable simulation modules, or constrained regression layers are promising directions.

\textbf{(3) Toward scientific grounding}: A shift from empirical optimization to physics-constrained learning is essential to achieve models that not only perform well statistically but also reflect the causality and mechanisms of Earth system processes.

\subsection{Generalization and Cross-Sensor Adaptability}

\textbf{(1) Cross-region transfer remains a bottleneck.} Inversion models trained on specific biomes, climatic zones, or land-use types often fail when applied to out-of-distribution regions, primarily due to shifts in surface properties, atmospheric conditions, and vegetation phenology \cite{peng2022domain, hu2025dualstrip}. This limits their application in global monitoring programs.

\textbf{(2) Sensor heterogeneity complicates data fusion.} Differences in spectral range, spatial resolution, radiometric calibration, and revisit frequency among sensors (e.g., Sentinel-2, Landsat-8, PlanetScope, MODIS) pose challenges to model generalization. Even models with strong multi-modal fusion mechanisms (e.g., GFM, mmEarth) may struggle to maintain consistent performance across modalities.

\textbf{(3) Future directions should emphasize:}

\begin{itemize}
    \item Developing domain-invariant feature representations that are robust to spatial, spectral, and temporal shifts.
    \item Leveraging unsupervised domain adaptation and contrastive alignment strategies for sensor harmonization.
    \item Constructing benchmark datasets explicitly designed for cross-sensor, cross-region, and multi-task evaluation.
\end{itemize}

\subsection{Data Constraints and Label Reliability}

\textbf{(1) High-quality ground truth remains scarce.} Many inversion targets—such as AGB, SIF, or soil moisture—require intensive field campaigns and are only sparsely sampled in space and time. This creates strong reliance on synthetic data or proxies, which may introduce systematic uncertainties.

\textbf{(2) Label quality often undermines training effectiveness.} Publicly available inversion datasets may suffer from temporal mismatch (e.g., satellite–field date misalignment), heterogeneous protocols (e.g., AGB plot scaling methods), or label noise due to propagation from upstream models. These issues can bias model learning and result in degraded physical fidelity.

\textbf{(3) Recommended future directions include:}

\begin{itemize}
    \item Using RTM-based simulations (e.g., SCOPE, DART) for weakly supervised or semi-supervised pretraining;
    \item Incorporating uncertainty quantification (e.g., Monte Carlo dropout, Bayesian ensembles) into regression pipelines;
    \item Applying label refinement through multi-source fusion, expert knowledge, or physically consistent filtering.
\end{itemize}

\textbf{In conclusion}, quantitative remote sensing inversion is evolving toward a cognition-driven modeling paradigm that fuses physical theory with data-centric learning. Advancing this transition will require:
(1) deeper integration of physical principles into model architectures,
(2) robust adaptation across regions, modalities, and platforms, and
(3) reliable and scalable data infrastructures grounded in scientific validation.
These efforts will collectively support the development of next-generation intelligent inversion systems capable of operating at continental to global scales for applications in climate monitoring, ecosystem assessment, and carbon cycle modeling.

\section{Conclusion}

Quantitative remote sensing inversion has evolved from physically grounded models to data-driven learning and, more recently, to foundation model–based paradigms that support multi-task, multi-modal estimation. Each approach offers distinct advantages and limitations in terms of interpretability, scalability, and generalization. While foundation models mark a promising step toward unified and flexible inversion frameworks, challenges remain in physical consistency, cross-domain adaptability, and data quality. Future efforts should prioritize the integration of physical principles, development of benchmark datasets, and collaboration across disciplines to build robust, interpretable, and globally deployable inversion systems.

\backmatter

% \bmhead{Supplementary information}

% If your article has accompanying supplementary file/s please state so here. 

% Authors reporting data from electrophoretic gels and blots should supply the full unprocessed scans for key as part of their Supplementary information. This may be requested by the editorial team/s if it is missing.

% Please refer to Journal-level guidance for any specific requirements.

% \bmhead{Acknowledgements}

% Acknowledgements are not compulsory. Where included they should be brief. Grant or contribution numbers may be acknowledged.

% Please refer to Journal-level guidance for any specific requirements.

\bmhead{Funding}
This study is supported by the National Natural Science Foundation of China (Grant No. 62466061); supported by the Yunnan Fundamental Research Projects (Grant No. 202401AU070052); supported by the Caizhou Technology Remote Sensing AI Innovation Research Fund (Grant No. CZ-RS-AI-2025-001); supported by the Yunnan Fundamental Research Projects
(No. 202501AT070456) and the Talent Introduction Project of Science Research Foundation of Yunnan University of Finance and Economics (No.2023D49).

\bmhead{Competing interests}
All authors have no conflict of interest.

\bmhead{Author contribution}
Zhenyu Yu: Conceptualization, Methodology, Writing – original draft, Writing – review \& editing. 
Mohd Yamani Inda Idris: Supervision, Writing – review \& editing. 
Hua Wang: Writing – review \& editing.
Pei Wang: Writing – review \& editing.
Junyi Chen: Writing – review \& editing. 
Kun Wang: Writing – review \& editing.
% Qi Wang: Writing – review \& editing.
% Fei Ma: Writing – review \& editing.
% Wenbin Zhang: Writing – review \& editing. 
% Rizwan Qureshi: Writing – review \& editing.
Xiang Yong: Writing – review \& editing.

% \section*{Declarations}

% Some journals require declarations to be submitted in a standardised format. Please check the Instructions for Authors of the journal to which you are submitting to see if you need to complete this section. If yes, your manuscript must contain the following sections under the heading `Declarations':

% \begin{itemize}
% \item Funding
% \item Conflict of interest/Competing interests (check journal-specific guidelines for which heading to use)
% \item Ethics approval and consent to participate
% \item Consent for publication
% \item Data availability 
% \item Materials availability
% \item Code availability 
% \item Author contribution
% \end{itemize}

\noindent
% If any of the sections are not relevant to your manuscript, please include the heading and write `Not applicable' for that section. 

%%===================================================%%
%% For presentation purpose, we have included        %%
%% \bigskip command. Please ignore this.             %%
%%===================================================%%
\bigskip
\bibliography{output}

\end{document}